\documentclass[11pt]{article}

\usepackage{epsfig}
\usepackage{a4}
\usepackage{nobi}
\usepackage{ims}
\usepackage{nobiDG}
\usepackage{colacl98}
\usepackage{nobibib}

\newcounter{defcnt}

\newcommand{\up}{\ensuremath{\uparrow}}
\newcommand{\down}{\ensuremath{\downarrow}}

\newenvironment{ruleset}{%
	\begin{center}
	\footnotesize
	\begin{tabular}{r@{}ccc}
}{%
	\end{tabular}
	\end{center}
}
\newcommand{\f}[1]{\textsc{#1}}

\begin{document}

\title{\appearedIn{\shortstack{Kahane, S. \& Polgu\`{e}re, A. (Eds, 1998): Proc. of COLING-ACL Workshop \\ 
			``Processing of Dependency-Based Grammars'', pages 29--38}%
	}{29}
	How to define a context-free backbone for DGs: \\
	Implementing a DG in the LFG formalism}

\author{Norbert Br\"{o}ker \\ \nobijoborga \\ \nobijobstrasse \\
	\nobijobplz\ \nobijobstadt \\ \textsc{\nobijobemail} }





\maketitle

\begin{abstract}
This paper presents a multidimensional Dependency Grammar (DG), which
decouples the dependency tree from word order, such that surface ordering
is not determined by traversing the dependency tree. We develop the notion
of a \emph{word order domain structure}, which is linked but structurally
dissimilar to the syntactic dependency tree. We then discuss the
implementation of such a DG using constructs from a unification-based
phrase-structure approach, namely Lexical-Functional Grammar (LFG).
Particular attention is given to the analysis of discontinuities in DG in
terms of LFG's functional uncertainty.
\end{abstract}

\section{Introduction
	\label{ch:intro}}
Recently, the concept of valency has gained considerable attention. Not
only do all linguistic theories refer to some reformulation of the
traditional notion of valency (in the form of $\theta$-grid,
subcategorization list, argument list, or extended domain of locality);
there is a growing number of parsers based on binary relations between
words \cite{Eisner1997,Maruyama1990}. Even theories based on phrase
structure may have processing models based on relations between lexical
items \cite{Rambow+Joshi-processing94}.

Against the background of this interest in the valency concept, and the
fact that word order is one of the main difference between phrase-structure
based approaches (henceforth PSG) and dependency grammar (DG), this paper
will propose a word order description for DG and describe its
implementation. First, we will motivate the separation of surface order and
dependency relations within DG, and make a specific architectural proposal
for their linking. Second, we will briefly sketch Lexical-Functional
Grammar (LFG), and then show in detail how one might use the formal
constructs provided by LFG to encode the proposed DG architecture.

Our position will be that dependency relations are motivated semantically
\cite{Tesniere1959}, and need not be projective. We argue for so-called
\emph{word order domains}, consisting of partially ordered sets of words
and associated with nodes in the dependency tree. These order domains
constitute a tree defined by set inclusion, and surface word order is
determined by traversing this tree. A syntactic analysis therefore consists
of two linked, but dissimilar trees.

The paper thus sheds light on two questions. A very early result on the
weak generative equivalence of context-free grammars and DGs suggested that
DGs are incapable of describing surface word order \cite{Gaifman1965}. This
result has been criticised to apply only to impoverished DGs which do not
properly represent formally the expressivity of contemporary DG variants
\cite{nobi:acl97}, and our use of a context-free backbone with further
constraints imposed by dependency relations further supports the view that
DG is not a notational variant of context-free grammar. The second question
addressed is that of efficient processing of discontinuous DGs. By
converting a native DG grammar into LFG rules, we are able to profit from
the state of the art in context-free parsing technology. A context-free
base (or skeleton) has often been cited as a prerequisite for practical
applicability of a natural language grammar \cite{Erbach1990}, and we here
show that a DG can meet this criterion with ease.

Sec.~\ref{ch:dg} will briefly review approaches to word order in DG, and
Sec.~\ref{ch:domains} introduces word order domains as our proposal. LFG is
briefly introduced in Sec.~\ref{ch:lfg}, and the encoding of DG within the
LFG framework is the topic of Sec.~\ref{ch:convert}.

\section{Word Order in DG
	\label{ch:dg}}

A very brief characterization of DG is that it recognizes only lexical, not
phrasal nodes, which are linked by directed, typed, binary relations to
form a dependency tree \cite{Tesniere1959,Hudson1993}. If these relations
are motivated semantically, such dependency trees can be non-projective.
Consider the extracted NP in \os{Beans, I know John likes}. A projective
tree would require \os{Beans} to be connected to either \os{I} or \os{know}
-- none of which is conceptually directly related to \os{Beans}. It is
\os{likes} that determines syntactic features of \os{Beans} and which
provides a semantic role for it. The only connection between \os{know} and
\os{Beans} is that the finite verb allows the extraction of \os{Beans},
thus defining order restrictions for the NP. The following overview of DG
flavors shows that various mechanisms (global rules, general graphs,
procedural means) are generally employed to lift the limitation of
projectivity and discusses some shortcomings of these proposals.

\paragraph{Functional Generative Description}
\cite{Sgall1986} assumes a language-independent \emph{underlying order},
which is represented as a projective dependency tree. This abstract
representation of the sentence is mapped via \emph{ordering rules} to the
concrete surface realization. Recently, \textcite{Kruijff1997} has given a
categorial-style formulation of these ordering rules. He assumes
associative categorial operators, permuting the arguments to yield the
surface ordering. One difference to our proposal is that we argue for a
representational account of word order (based on valid structures
representing word order), eschewing the non-determinism introduced by unary
categorial operators; the second difference is the avoidance of an
underlying structure, which stratifies the theory and makes incremental
processing difficult.

\paragraph{Meaning-Text Theory}
\cite{Melcuk1988} assumes seven \emph{strata of representation}. The rules
mapping from the unordered dependency trees of surface-syntactic
representations onto the annotated lexeme sequences of deep-morphological
representations include \emph{global ordering rules} which allow
discontinuities. These rules have not yet been formally specified
\cite[p.187f]{Melcuk+Pertsov1987} (but see the proposal by
\textcite{Rambow+Joshi-MTT}).

\paragraph{Word Grammar}
(WG, \textcite{Hudson1990}) is based on \emph{general graphs} instead of
trees. The ordering of two linked words is specified together with their
dependency relation, as in the proposition ``\texttt{object of verb follows
it}''. Extraction of, e.g., objects is analyzed by establishing an
additional dependency called \texttt{visitor} between the verb and the
extractee, which requires the reverse order, as in ``\texttt{visitor of
verb precedes it}''. Resulting inconsistencies, e.g. in case of an
extracted object, are not resolved. This approach compromises the semantic
motivation of dependencies by adding \emph{purely order-induced
dependencies}.

\paragraph{Dependency Unification Grammar}
(DUG, \textcite{Hellwig1986}) defines a tree-like data structure for the
representation of syntactic analyses. Using morphosyntactic features with
special interpretations, a word defines \emph{abstract positions} into
which modifiers are mapped. Partial orderings and even discontinuities can
thus be described by allowing a modifier to occupy a position defined by
some transitive head. The approach requires that the \emph{parser}
interprets several features in a special way, and it cannot restrict the
scope of discontinuities.

\paragraph{Slot Grammar}
\cite{McCord1990} employs a number of rule types, some of which are
exclusively concerned with precedence. So-called head/slot and slot/slot
\emph{ordering rules} describe the precedence in projective trees,
referring to arbitrary predicates over head and modifiers. Extractions
(i.e., discontinuities) are merely handled by a mechanism built into the
\emph{parser}.

\section{Word Order Domains
	\label{ch:domains}}

Extending the previous discussion, we require the following of a word order
description for DG:

\begin{itemize}
\item not to compromise the semantic motivation of dependencies,
\item to be able to restrict discontinuities to certain constructions and
delimit their scope,
\item to be lexicalized without requiring lexical ambiguities for the
representation of ordering alternatives,
\item to be declarative (i.e., independent of an analysis procedure), and
\item to be formally precise and consistent.
\end{itemize}

The subsequent definition of an order domain structure and its linking to
the dependency tree satisify these requirements.

\subsection{The Order Domain Structure}

A \emph{word order domain} is a set of words, generalizing the notion of
positions in DUG. The cardinality of an order domain may be restricted to
at most one element, at least one element, or -- by conjunction -- to
exactly one element. Each word is associated with a sequence of order
domains, one of which must contain the word itself, and each of these
domains may require that its elements have certain features. Order domains
can be partially ordered based on set inclusion: If an order domain $d$
contains word $w$ (which is not associated with $d$), every word $w'$
contained in a domain $d'$ associated with $w$ is also contained in $d$;
therefore, $d' \subset d$ for each $d'$ associated with $w$. This partial
ordering induces a tree on order domains, which we call the \emph{order
domain structure}. The order domain structure constitutes a projective tree
over the input, where order domains loosely correspond to partial phrases.

\begin{figure}[t]
\centering
\epsfig{file=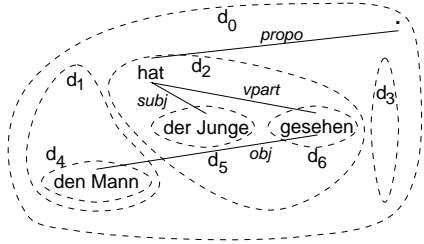,width=0.7\columnwidth}
\caption{Dependency Tree and Order Domains for \ref{ex:1}
	\label{fig:dep-tree}}
\end{figure}

\begin{figure}[t]
\centering
\epsfig{file=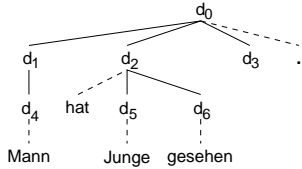,width=0.5\columnwidth}
\caption{Order Domain Structure for \ref{ex:1}
	\label{fig:dom-tree}}
\end{figure}

\ex{\label{ex:1}
\gloss{4}%
	{Den Mann & hat & der Junge & gesehen.}%
	{the man$_{ACC}$ & has & the boy$_{NOM}$ & seen}%
	{The boy has seen the man.}}

\noindent
Take the German example \ref{ex:1}. Its dependency tree is shown in
Fig.~\ref{fig:dep-tree}, with word order domains indicated by dashed
circles. The finite verb, \os{hat}, defines a sequence of domains,
\tuple{d_1, d_2, d_3}, which roughly correspond to the topological fields
in the German main clause. The nouns and the participle each define a
single order domain. Set inclusion gives rise to the domain structure in
Fig.~\ref{fig:dom-tree}, where the individual words are attached by dashed
lines to their including domains.

\subsection{Surface Ordering}

How is the surface order derived from an order domain structure? First of
all, the ordering of domains is inherited by their respective elements,
i.e., \os{Mann} precedes (any element of) $d_2$, \os{hat} follows (any
element of) $d_1$, etc.

Ordering within a domain, e.g., of \os{hat} and $d_6$, or $d_5$ and $d_6$,
is based on precedence predicates (adapting the precedence predicates of
WG). There are two different types, one ordering a word with respect to any
other element of the domain it is associated with (e.g., \os{hat} with
respect to $d_6$), and another ordering two modifiers, referring to the
dependency relations they occupy ($d_5$ and $d_6$, referring to
\texttt{subj} and \texttt{vpart}). A verb like \os{hat} introduces three
precedence predicates, requiring other words (within the same domain) to
follow itself and the participle to follow subject and object, resp.:%
\footnote{%
For more details on the exact syntax and the semantics of these
propositions, see \cite{nobi:coling98}. 
}

\begin{center}
\begin{tabular}{rl}
\os{hat} \IMPL 	& $<_{*}$  \\
	\AND	& \texttt{subj} $<$ \texttt{vpart} \\
	\AND	& \texttt{obj} $<$ \texttt{vpart}
\end{tabular}
\end{center}

Informally, the first conjunct is satisfied by any domain in which no word
precedes \os{hat}, and the second and third conjuncts are satisfied by any
domain in which no \texttt{subj}ect or \texttt{obj}ect follows a participle
(\texttt{vpart}). The \texttt{obj} must be mentioned for \os{hat}, although
\os{hat} does not directly govern objects, because objects may be placed by
\os{hat}, and not their immediate governors. The domain structure in
Fig.\ref{fig:dom-tree} satisfies these restrictions since nothing follows
the participle, and because \os{den Mann} is not an element of $d_2$, which
contains \os{hat}. This is an important interaction of order domains and
precedence predicates: Order domains define scopes for precedence
predicates. In this way, we take into account that dependency trees are
flatter than PS-based ones%
\footnote{
Note that each phrasal level in PS-based trees defines a scope for linear
precedence rules, which only apply to sister nodes. 
}
and avoid the formal inconsistencies noted above for WG.

\subsection{Linking Domain Structure and Dependency Tree}

Order domains easily extend to discontinuous dependencies. Consider the
non-projective tree in Fig.\ref{fig:dep-tree}. Assuming that the finite
verb governs the participle, no projective dependency between the object
\os{den Mann} and the participle \os{gesehen} can be established. We allow
non-projectivity by loosening the linking between dependency tree and
domain structure: A modifier (e.g., \os{Mann}) may not only be inserted
into a domain associated with its direct head (\os{gesehen}), but also into
a domain of a transitive head (\os{hat}), which we will call the
\emph{positional head}.

The possibility of inserting a word into a domain of some transitive head
raises the questions of how to require continuity (as needed in most
cases), and how to limit the distance between the governor and the
modifier. Both questions will be solved with reference to the dependency
relation. From a descriptive viewpoint, the \emph{syntactic construction}
is often cited to determine the possibility and scope of discontinuities
\cite{Bhatt1990,Matthews1981}. In PS-based accounts, the construction is
represented by phrasal categories, and extraction is limited by bounding
nodes (e.g., \textcite{Haegeman1994,Becker1991}). In dependency-based
accounts, the construction is represented by the dependency relation, which
is typed or labelled to indicate constructional distinctions which are
configurationally defined in PSG. Given this correspondence, it is natural
to employ dependencies in the description of discontinuities as follows:
For each modifier, a set of dependency types is defined which may link the
direct head and the positional head of the modifier (\os{gesehen} and
\os{hat}, respectively). If this set is empty, both heads are identical and
a continuous attachment results. The impossibility of extraction from,
e.g., a finite verb phrase follows from the fact that the dependency
embedding finite verbs, \texttt{propo}, may not appear on any path between
a direct and a positional head.

\section{A Brief Review of LFG
	\label{ch:lfg}}

This section introduces key concepts of LFG which are of interest in
Sec.~\ref{ch:convert} and is necessarily very short. Further information can
be found in \ct{Bresnan1982} and \ct{Dalrymple-etal1995}.

LFG posits several different representation levels, called
\emph{projections}. Within a projection, a certain type of linguistic
knowledge is represented, which explains differences in the formal setup
(data types and operations) of the projections. The two standard
projections, and those used here, are the constituent (c-) structure and
the functional (f-) structure (\ct{Kaplan1995} and \ct{Halvorsen1995}
discuss the projection idea in more detail). C-structure is defined in
terms of context-free phrase structure rules, and thus forms a projective
tree of categories over the input. It is assumed to encode language
particularities with respect to the set of categories and the possible
orderings. The f-structure is constructed from additional annotations
attached to the phrase structure rules, and has the form of an
attribute-value matrix or feature structure. It is assumed to represent
more or less language-independent information about grammatical functions
and predicate-argument structure. In addition to the usual unification
operation, LFG employs existential and negative constraints on features,
which allow the formulation of constraints about the existence of features
without specifying the associated value.

\begin{figure*}[tb]
\centering
\epsfig{file=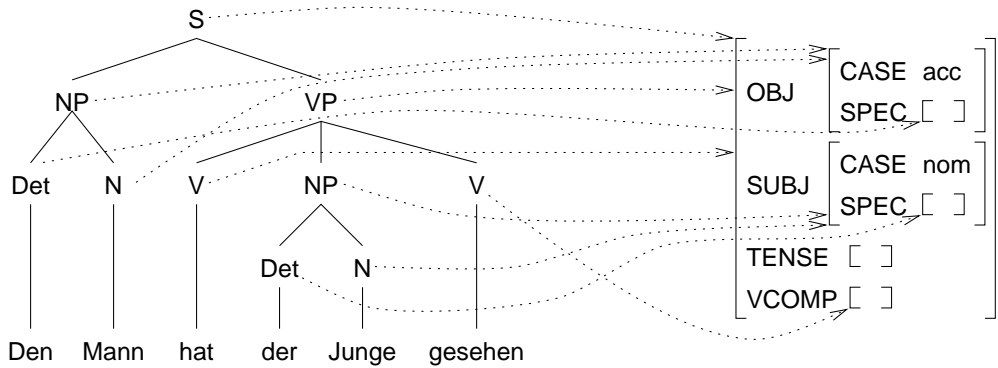,width=0.8\textwidth}
\caption{C-structure (left) and f-structure (right) for \ref{ex:1}
	\label{fig:c-f-struc}}
\end{figure*}

Consider the following rules, which are used for illustration only and do
not constitute a canonical LFG analysis.

\begin{ruleset}
S \prod		& NP & VP \\
	& (\up \f{obj})=\down
		& \up =\down \\
	& (\down \f{case})=acc \\
& \\
NP \prod	& Det & N \\
	& (\up \f{spec})=\down 
		& \up =\down \\
& \\
VP \prod	& V & NP & V \\
	& \up =\down 
		& (\up \f{subj})=\down 
			& (\up \f{vcomp})=\down \\
	& (\down \f{tense})
		& (\down \f{case})=nom
			& $\sim$(\down \f{tense}) \\
\end{ruleset}

Assuming reasonable lexical insertion rules, the context-free part of these
rules assigns the c-structure to the left of Fig.~\ref{fig:c-f-struc} to
example \ref{ex:1}. The annotations are associated with right-hand side
elements of the rules and define the f-structure of the sentence, which is
displayed to the right of Fig.~\ref{fig:c-f-struc}. Each c-structure node
is associated with an f-structure node as shown by the arrows. The
f-structure node associated with the left-hand side of a rule may be
accessed with the \up\ metavariable, while the f-structure node of a
right-hand side element may be accessed with the \down\ metavariable. The
mapping from c-structure nodes to f-structure nodes is not one-to-one,
however, since the feature structures of two distinct c-structure nodes may
be identified (via the \up=\down\ annotation), and additional embedded
features may be introduced (such as CASE). Assuming that only finite verbs
carry the TENSE feature, the existential constraint (\down TENSE) requires
a finite verb at the beginning of the VP, while the negative constraint
$\sim$(\down TENSE) forbids finite verbs at the end of the VP. Note that
unspecified feature structures are displayed as [ ] in the figure, and that
much more information (esp. predicate-argument information) will come from
the lexical entries.

Another important construct of LFG is functional uncertainty
\cite{Kaplan+Zaenen1995,Kaplan+Maxwell1995}. Very often (most notably, in
extraction or control constructions) the path of f-structure attributes to
write down is indeterminate. In this case, one may write down a description
of this path (using a regular language over attribute names) and let the
parser check every path described (possibly resulting in ambiguities
warranted by f-structure differences only). Our little grammar may be
extended to take advantage of functional uncertainty in two ways. First, if
you want to permute subject and object (as is possible in German), you
might change the S rule to the following:

\begin{ruleset}
S \prod		& NP & VP \\
	& (\up \{\f{obj} \| \f{subj}\})=\down
		& \up =\down \\
\end{ruleset}

The f-structure node of the initial NP may now be inserted in either the
OBJ or the SUBJ attribute of the sentence's f-structure, which is expressed
by the disjunction \{OBJ\|SUBJ\} in the annotation. (Of course, you have to
restrict the CASE feature suitably, which can be done in the verb's
subcategorization.) The other regular notation which we will use is the
Kleene star. Assume a different f-structure analysis, where the object of
infinite verbs is embedded under VCOMP. The S rule from above would have to
be changed to the following:

\begin{ruleset}
S \prod		& NP & VP \\
	& (\up \{(\f{vcomp}) \f{obj} \| \f{subj}\})=\down
		& \up =\down \\
\end{ruleset}

But this rule will only analyse verb groups with zero or one auxiliary,
because the VCOMP attribute is optional in the path description. Examples
like \emph{Den Mann will der Junge gesehen haben} with several auxiliaries
are not covered, because the main verb is embedded under (VCOMP VCOMP). The
natural solution is to use the Kleene star as follows, which allows zero or
more occurrences of the attribute VCOMP.

\begin{ruleset}
S \prod		& NP & VP \\
	& (\up \{\f{vcomp}* \f{obj} \| \f{subj}\})=\down
		& \up =\down \\
\end{ruleset}

A property which is important for our use of functional uncertainty is
already evident from these examples: Functional uncertainty is
non-constructive, i.e., the attribute paths derived from such an annotation
are not constructed anew (which in case of the Kleene star would lead to
infinitely many solutions), but must already exist in the f-structure.

\section{Encoding DG in LFG
	\label{ch:convert}}

\subsection{The Implementation Plattform
	\label{ch:xle}}

The plattform used is the Xerox Linguistic Environment (XLE, see also
\texttt{\small http://www.parc.xerox.com/istl/groups/nltt/xle/}), which
implements a large part of LFG theory plus a number of abbreviatory
devices. It includes a parser, a generator, support for two-level
morphology and different types of lexica as well as a user-friendly
graphical interface with the ability to browse through the set of analyses,
to work in batch mode for testing purposes, etc.

We will be using two abbreviatory devices below, which are shortly
introduced here. Both do not show up in the final output, rather they allow
the grammar writer to state various generalizations more succintly. The
first is the so-called \emph{metacategory}, which allows several
c-structure categories to be merged into one. So if we are writing
\ref{ex:metacat}, we introduce a metacategory \texttt{domVfin}
(representing the domain sequence of finite verbs) to be used in other
rules, but we will never see such a category in the c-structure. Rather,
the expansion of the metacategory is directly attached to the mother node
of the metacategory (cf. Fig.~\ref{fig:c-struc-window}).

\ex{\label{ex:metacat}
\small\tt
	domVfin~=~domINITIAL~domMIDDLE~domFINAL.}

The second abbreviatory construct is the \emph{template}, which groups
several functional annotations under one heading, possibly with some
parameters. A very important template is the \texttt{VALENCY} template
defined in \ref{ex:temp-def}, which defines a dependency relation on
f-structure (see below for discussion). We require three parameters (each
introduced by underscore), the first of which indicates optionality
(\texttt{opt} vs. \texttt{req} values), the second gives the name of the
dependency relation, and the third the word class required of the modifier.
\ref{ex:temp-use} shows a usage of a template, which begins with an @ (at)
sign and lists the template name with any parameters enclosed in
parentheses.

\ex{\label{ex:temp-def}
\small\tt
\begin{tabular}{@{}rl}
VALENCY	(\_o \_d \_c) = \{ 
	& \_o = opt \\
	& $\sim$(\up \_d) \\
\|	& (\up \_d CLASS) = \_c \\
	& (\up \_d LEXEME) \}.
\end{tabular}
}
\ex{\label{ex:temp-use}
\small\tt
@(VALENCY req OBJ N).
}

\subsection{Topological fields
	\label{ch:fields}}

\begin{figure}
\centering
\epsfig{file=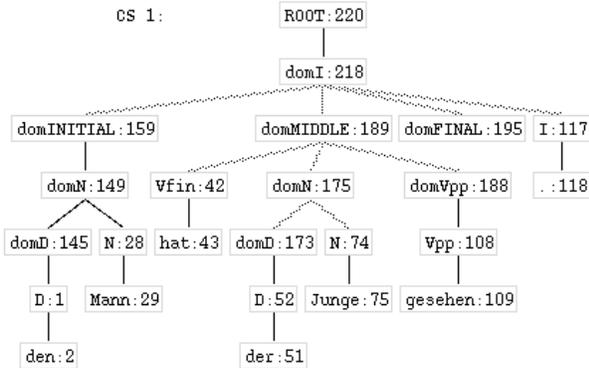,width=\columnwidth}
\caption{C-structure for \ref{ex:1}
	\label{fig:c-struc-window}}
\end{figure}

As we have seen in Sec.~\ref{ch:domains}, the order domain structure is a
projective tree over the input. So it is natural to encode the domain
structure in context-free rules, resulting in a tree as shown in
Fig.~\ref{fig:c-struc-window}. Categories which have a status as order
domains are named \texttt{dom*}, to be distinguishable from preterminal
categories (such as \texttt{Vfin}, \texttt{I}, \ldots; these cannot be
converted to metacategories). As notational convention, \texttt{domC} will
be the name of the (meta)category defining the order domain sequence for a
word of class \texttt{C}. Eliminating the preterminal categories yields
exactly the domain structure given in Fig.~\ref{fig:dom-tree}.

A complete algorithmic description of how to derive phrase-structure rules
from order domain definitions would require a lenghty introduction to more
of XLE's c-structure constructs, and therefore we illustrate the conversion
with handcoded rules. For example, a noun introduces one order domain
without cardinality restrictions. Assuming a metacategory \texttt{DOMAIN}
standing for an arbitrary domain, we define the following rules for the
domain sequences of nouns, full stops, and determiners:

\ex{
\small\tt
	domN \prod\ DOMAIN* N DOMAIN*. \\
	domI \prod\ DOMAIN I. \\
	domD \prod\ D.}

A complex example is the finite verb, which introduces three domains, each
with different cardinality restrictions. This is encoded in the following
rules: 

\ex{
\small\tt
\begin{tabular}{@{}l@{ }l}
domVfin		& = domINITIAL domMIDDLE domFINAL. \\
domINITIAL	& \prod\ DOMAIN. \\
domMIDDLE	& \prod\ DOMAIN* Vfin DOMAIN*. \\
domFINAL	& \prod\ ( DOMAIN ). 
\end{tabular}
}

Note the use of a metacategory here, which does not appear in the
c-structure output (as seen in Fig.~\ref{fig:c-struc-window}), but still
allows you to refer to all elements placed by a finite verb in one word.
The definition of \texttt{DOMAIN} is trivial: It is just a metacategory
expandable to every domain:%
\footnote{%
A number of efficiency optimizations can be directly compiled into these
c-structure rules. Mentioning \texttt{DOMAIN} is much too permissive in
most cases (e.g., within the NP), and can be optimized to allow only
domains introduced by words which may actually be modifiers at this
point. 
}

\ex{
\small\tt
	DOMAIN = \{ domVfin | domI | domN | domD \}.
}

\subsection{Valencies and Dependency Relations
	\label{ch:deprel}}

The dependency tree is, at least in our approach, an unordered tree with
labelled relations between nodes representing words. This is very similar
to the formal properties of the f-structure, which we will therefore use to
encode it. We have already presented the \texttt{VALENCY} template in
\ref{ex:temp-def} and will now explain it.  \texttt{$\{\cdots\|\cdots\}$}
represents a disjunction of possibilities, and the parameter \texttt{\_o}
(for optionality) controls their selection. In case we provide the
\texttt{opt} value, there is an option to forbid the existence of the
dependency, expressed by the negative constraint \texttt{$\sim$(\up \_d)}.
Regardless of the value of \texttt{\_o}, there is another option to
introduce an attribute named \texttt{\_d} (for dependency) which contains a
\texttt{CLASS} attribute with a value specified by the third parameter,
\texttt{\_c}.  The existential constraint for the \texttt{LEXEME} attribute
requires that some other word (which specifies a \texttt{LEXEME}) is
unified into the feature \texttt{\_d}, thereby filling this valency
slot. The use of a defining constraint for the \texttt{CLASS} attribute
constructs the feature, allowing non-constructive functional uncertainty to
fill in the modifier (as explained below). 

A typical lexical entry is shown in \ref{ex:lex}, where the surface form is
followed by the c-structure category and some template invocations. These
expand to annotations defining the \texttt{CLASS} and \texttt{LEXEME}
features, and use the \texttt{VALENCY} template to define the valency
frame. 

\ex{\label{ex:lex}
\small\tt
\begin{tabular}{ll}
hat Vfin 	& @(Vfin aux-perfect\_) \\
		& @(VALENCY req SUBJ N) \\
		& @(VALENCY req VPART Vpp). 
\end{tabular}
}

\subsection{Continuous and Discontinuous Attachment
	\label{ch:disco}}

\begin{figure*}
\centering
\epsfig{file=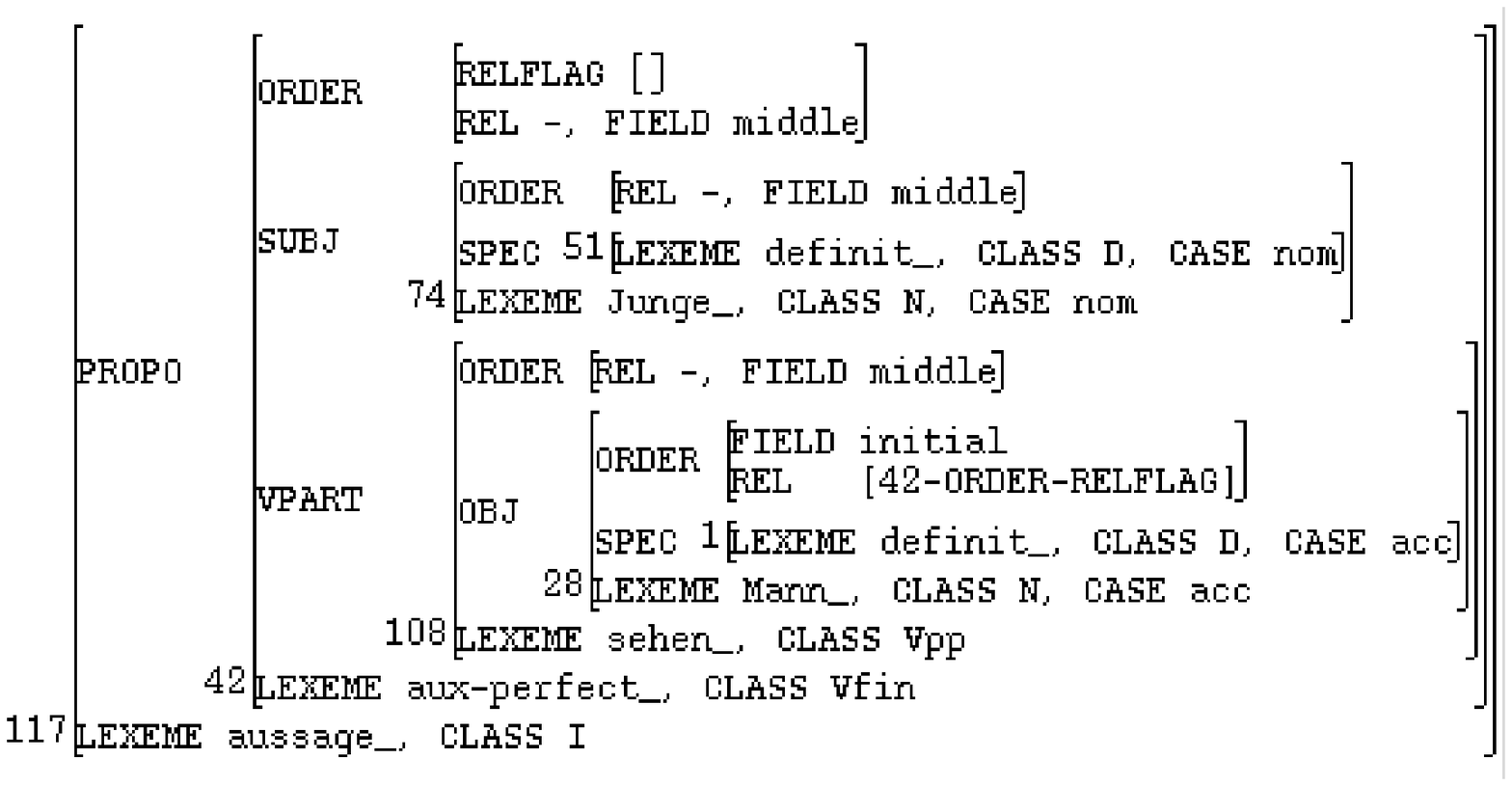,width=0.7\textwidth}
\caption{F-structure for \ref{ex:1}
	\label{fig:f-struc-window}}
\end{figure*}

So far we get only a c-structure where words are associated with
f-structures containing valency frames. To get the f-structure shown in
Fig.~\ref{fig:f-struc-window} (numbers refer to c-structure node numbers of
Fig.~\ref{fig:c-struc-window}) we need to establish dependency relations,
i.e., need to put the f-structures associated with preterminal nodes
together into one large f-structure. Establishing dependency relations
between the words relies heavily on the mechanism of functional
uncertainty. First, we must identify on f-structure the head of each order
domain sequence. For this, we annotate in every c-structure rule the
category of the head word with the template \texttt{@(HEAD)}, which
identifies the head word's f-structure with the order domain's f-structure
(cf. \ref{ex:head}). Second, all other c-structure categories (which
represent modifiers) are annotated with the \texttt{@(MODIFIER)} template
defined in \ref{ex:mod}. This template states that the f-structure of the
modifier (referenced by \down) may be placed under some dependency
attribute path of the f-structure of the head (referenced by \up). These
paths are of the form \texttt{p d}, where \texttt{p} is a (possibly empty)
regular expression over dependency attributes, and \texttt{d} is a
dependency attribute. \texttt{d} names the dependency relation the modifier
finally fills, while \texttt{p} describes the path of dependencies which
may separate the positional from the direct head of the modifier. The
\texttt{MODIFIER} template thus completely describes the legal
discontinuities: If \texttt{p} is empty for a dependency \texttt{d},
modifiers in dependency \texttt{d} are always continuously attached (i.e.,
in an order domain defined by their direct head). This is the case for the
subject (in dependency \texttt{SUBJ}) and the determiner (in dependency
\texttt{SPEC}), in this example. On the other hand, a non-empty path
\texttt{p} allows the modifier to `float up' the dependency tree to any
transitive head reachable via \texttt{p}. In our example, objects depending
on participles may thus float into domains of the finite verb (across
\texttt{VPART} dependencies), and relative clauses (in dependency
\texttt{RELA}) may float from the noun's domain into the finite verb's
domains.

\ex{\label{ex:head}
	\tt\small
	HEAD = \down =\up. 
}
\ex{\label{ex:mod}
	\tt\small
\begin{tabular}{@{}r@{}l}
MODIFIER = \down =(\up \{
	& PROPO \\
|	& SUBJ \\
|	& VPART* OBJ \\
|	& VPART \\
|	& SPEC \\
|	& \{SUBJ\|OBJ\|VPART\}* RELA\}).
\end{tabular}
}

The grammar defined so far overgenerates in that, e.g., relative clauses
may be placed into the middle field. To require placement in specific
domains, additional features are used, which distinguish topological fields
(e.g., via \texttt{(\down FIELD) = middle} annotations on c-structure). A
relative clause can then be constrained to occur only in the final field by
adding constraints on these features. This mechanism is very similar to
describing agreement or government (e.g., of case or number), which also
uses standard features not discussed here. With these additions, the final
rules for finite verbs look as follows:

\ex{
\small\tt\hspace{-1em}
\begin{tabular}{@{}l@{}r@{}l}
domINITIAL \prod\ 
	& DOMAIN:
		& @(MODIFIER) \\
	&	& (\down FIELD) = initial. \\
domMIDDLE \prod\  
	& Vfin: 
		& @(HEAD) \\
	& 	& (\down FIELD) = middle; \\
	& DOMAIN*:
		& @(MODIFIER) \\ 
	& 	& (\down FIELD) = middle; \\
domFINAL \prod 
	& ( DOMAIN: 
		& @(MODIFIER) \\
	&	& (\down FIELD) = final ).
\end{tabular}
}

\subsection{Missing Links
	\label{ch:missing}}

As is to be expected if you use something for purposes it was not designed
to be used for, there are some missing links. The most prominent one is the
lack of binary precedence predicates over dependency relations. There is,
however, a close relative, which might be used for implementing precedence
predicates. \ct{Zaenen+Kaplan1995} introduced \emph{f-precedence} $<_f$
into LFG, which allows to express on f-structure constraints on the order
of the c-structure nodes mapping to the current f-structure. So we might
write the following annotations to order the finite verb with respect to
its modifiers, or to order subject and object.

\ex{
\small\tt
	(\up) $<_f$ (\up \{SUBJ\|OBJ\|VPART\}). \\
	(\up SUBJ) $<_f$ (\up OBJ).}

The problem with f-precedence, however, is that is does not respect the
scope restrictions which we defined for precedence predicates. I.e., a
topicalized object is not exempt from the above constraints, and thus would
result in parsing failure. To restrict the scope of f-precedence to order
domains (aka, certain c-structure categories) would require an explicit
encoding of these domains on f-structure.

\section{Conclusion
	\label{ch:conclu}}

We have presented a new approach to word order which preserves traditional
notions (semantically motivated dependencies, topological fields) while
being fully lexicalized and formally precise \cite{nobi:diss}. Word order
domains are sets of partially ordered words associated with words. A word
is contained in an order domain of its head, or may float into an order
domain of a transitive head, resulting in a discontinuous dependency tree
while retaining a projective order domain structure. Restrictions on the
floating are expressed in a lexicalized fashion in terms of dependency
relations. We have also shown how the order domains can be used to define a
context-free backbone for DG, and used a grammar development environment
for annotated phrase-structure grammars to encode the DG.

A number of questions immediately arise, some of which will hopefully be
answered until the time of the workshop. On the theoretical side, this work
has argued for a strict separation of precedence and categorial information
in LFG (or PSG in general, see \cite{nobi:lfg98}). Can these analyses and
insights be transferred? On the practical side, can the conversion we
sketched be used to create efficient large-scale DGs? Or will the amount of
f-structural indeterminacy introduced by our use of functional uncertainty
lead to overly long processing? And, last and most challenging, when will
the first large treebank with dependency annotation be available, and will
it be derived from XLE's f-structure output?

\smallbibliography{\small}
\bibliography{ref,nobi,add-on}
\bibliographystyle{nobibib}

\end{document}